# Iris segmentation techniques to recognize the behavior of a vigilant driver

Abdullatif BABA
Mechatronics Engineering department
University of Turkish Aeronautical Association, Ankara, Turkey
ababa@thk.edu.tr

*Abstract*— **In this paper, we clarify how to recognize different levels of vigilance for vehicle drivers. In order to avoid the classical problems of crisp logic, we preferred to employ a fuzzy logic-based system that depends on two variables to make the final decision. Two iris segmentation techniques are well illustrated. A new technique for pupil position detection is also provided here with the possibility to correct the pupil detected position when dealing with some noisy cases.**

## I. INTRODUCTION

A vehicle driver has to keep awake while driving in order to avoid any accident; it is a common collaborative mission for the road society. The main goal here is to detect the moments during which the driver is not vigilant enough to continue driving. Therefore, he has to be alerted by a smart system that is installed in his car. Martin et al. [11] analyzed the driver's gaze dynamic patterns for maneuvering freeway driving mode. The main objective here is to detect the most important zones of driver interest. Some other papers [7,2,6,10] suppose that the control of the vehicle is shared partially or completely with a robot system.

In this paper, we don't analyze the driver's gaze as it is already achieved in [9, 12, 3, 4]. But we present a visual-based alert system which is composed of two monoscopic cameras mounted at two symmetric points at a virtual arc in front of the driver as it is illustrated in "fig. 1". In this case, the driver's eyes are completely seen and always detected at each moment from any given direction.

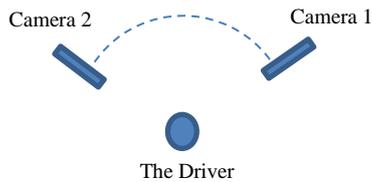

Fig. (1), The general design of our system, composed of two cameras mounted at two symmetric points in front of the driver.

## II. PUPIL SEGMENTATION AND PUPIL CENTER CORRECTION

Before explaining pupil detection and iris segmentation, face recognition and eye localization should be achieved here as a first step. In this work, face recognition is done using a method which was already published in our article "Face recognition with illumination varying conditions and occlusion" [1], where we used an approach that depends on some robust preprocessing techniques like PCA and LDA to improve the capability of features extraction when a small number of training samples is available and to deal with samples of occluded features. Correlation classifier was also used to reduce the number of misclassification states. Human eye localization will be done here using Haar wavelet-based technique [9].

The second step now is to detect the eye's pupil. In order to magnify the differences between neighbored pixels belonging to different textures like pupil, iris, or eyelid, the second derivative of Gaussian, described in equation 1 and illustrated in the 3D mesh grid in "fig. 2", is applied as a band-pass filter (in the frequency domain) to the input image.

$$H(u,v) = \frac{1}{\sigma^2}\left(\frac{D^2}{D_0^2} - 2\right) e^{-\frac{D^2}{2D_0^2}} \quad (1)$$

$D_0$ represents the pre-selected band of frequencies.

$D = \sqrt{u^2 + v^2}$; $\sigma$ is the standard deviation of the Gaussian distribution

As a result of this step, the pupil pixels will be recognized as a group of the highest intensities (peaks) comparing with their neighborhood as it is shown in "fig. 3". To find the center of this group of pixels, a template of odd size (31 x 31) scans the entire image to produce a new image where the intensity of each pixel is accumulated of its own intensity and the intensities of all neighbors presented by the template. In this case, the maximal accumulated value will be assigned to a unique pixel (the summit) which represents the center of the pupil; as it is illustrated in "fig. 4".

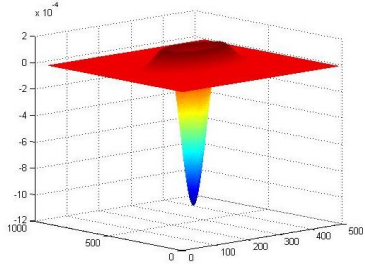

Fig. (2), The 3D mesh of the second derivative of the Gaussian distribution

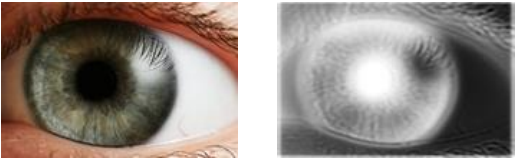

Fig. (3). The iris image, before and after filtering using (Frequency-built Band Pass Filter)

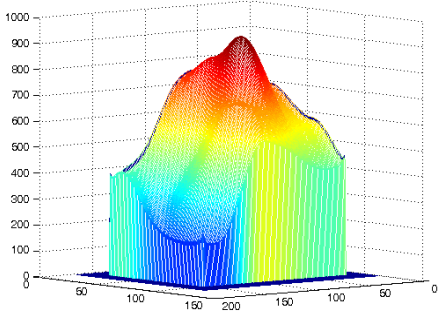

Fig. (4), The pupil center illustrated as a summit where the accumulated intensity is maximal.

Sometimes, the flashlight of the used camera may appear as a white spot located on the iris region. In some other images, the closest pixels around the pupil could be corrupted by noise. In both cases, the position of the pupil center which has been detected using the upper-mentioned technique becomes incorrect. To re-correct its location, an additional step should be done as it is clarified in "fig. 5". In this example, from the incorrect position of the pupil center, two virtual lines (horizontal and vertical) should be created without exceeding the limits of the pupil circle. The horizontal line hits the pupil circle in two points (P-right and P-left) while the second line hits the circle in two different vertical points (P-up and P-down). According to the Cartesian reference shown in "fig. 5"; the corrected coordinates of the pupil center are given as follows:

$$x_c = x_{P_{left}} + \left(x_{P_{right}} - x_{P_{left}}\right)/2 \quad (2)$$
$$y_c = y_{P_{up}} + \left(y_{P_{down}} - y_{P_{up}}\right)/2 \quad (3)$$

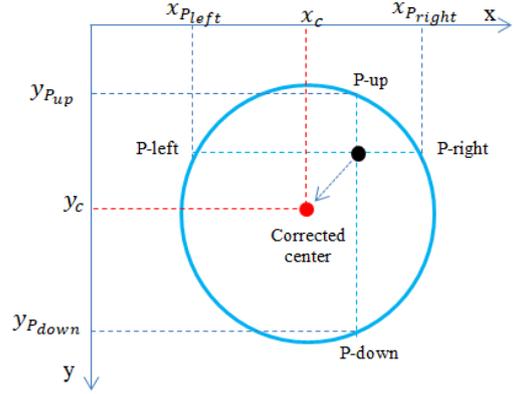

Fig. (5), Relocate the pupil center into its correct position

### III. IRIS SEGMENTATION

This step will be achieved by using two different techniques; the entropy-based texture description and the combination of Gabor filter with PCA.

#### A. Entropy-based texture description

The idea here is to generate a single ripple with an incremental radius (pixel by pixel) starting from the boundaries of the detected pupil, "fig. 6". To collect the spectral data for each new considered region; Fourier transformation will be calculated for each group of iris pixels coincide with each recently extended circle.

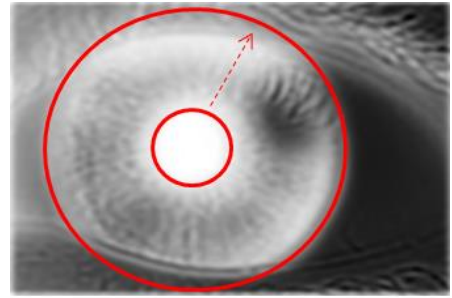

Fig. (6), A single ripple with an incremental radius (pixel by pixel) starting from the boundaries of the detected pupil

The entropy for each generated circle will be calculated as following [8]:

$$h = \sum_{u=1}^{N} \sum_{v=1}^{M} NFC_{u,v} \log(NFC_{u,v}) \quad (4)$$

u and v; are the conventional variables of the produced frequency rectangle on rows and columns respectively.

N and M: are the number of samples on rows and columns respectively

$NFC_{u,v}$; normalized Fourier coefficient which is calculated as following [8]:

$$NFC_{u,v} = \frac{|F_{u,v}|}{\sqrt{\sum_{u,v} F_{u,v}^2}} \quad (5)$$

$F_{u,v}$: is the spectral data of each pixel in the frequency rectangle after achieving Fourier transformation.

Within the iris region that has a coherent texture, if the calculated error (e) for two successive circles doesn't exceed a predefined maximal threshold ($e_{max}$); then the actual circle is included inside the iris region. Otherwise, the process stops, and the actual circle is the external boundaries of the iris region, "fig. 7".

$$e = h_{i+1} - h_i \quad (6)$$

h: is the calculated entropy in equation 4.

i: is the order of the successively generated circle.

B. Gabor filter-based technique:

To determine its impulse response, a Gaussian envelope function has to be multiplied by a complex sinusoid carrier. According to this concept; Gabor filter is widely used as a bandpass filter for different applications (features extraction, texture analysis, fingerprints recognition, stereo disparity estimation, face recognition ...etc.).

In this paper we use an array of Gabor filter which is tuned to several wavelengths and orientations, where we have selected four orientation samples in steps of 45 degrees, while wavelength samples are growing up from an experimental minimal value to another maximal value; both of them are respectively given as follow:

$$WaveLengthMin = 2/\sqrt{3} \quad (7)$$

$$WaveLengthMax = sqrt(ImgRows^2 + ImgCols^2). \quad (8)$$

ImgRows and ImgCols; are respectively the number of rows and columns of the input image.

In this experiment; Gabor filter is used with the famous statistical procedure PCA (Principal Component Analysis) in order to construct Gabor feature sets to distinguish between iris pixels and the other pixels. According to this principle; we reshape the features sets to an acceptable form to be read by the PCA; to do that we add some spatial location information in both X and Y to create groups of features that are spatially close together. For example, if there are 16 Gabor features (4 wavelength & 4 orientations) we add 2 spatial features in X and Y for each pixel, that means, the size of the variable "featureSet" will be finally [M x N x 18] as it is illustrated in "fig. 8". Then the features are normalized to be zero mean with unit variance. Now, PCA determines the principal component coefficients for the M*N data matrix. The size of the matrix determined by the PCA in our example is [18*18]. Each column of this matrix, shown in "fig. 9", contains coefficients for one principal component.

For the same original image which was already manipulated and illustrated in "fig. 7", if we apply the Gabor based technique, we get the result shown in "fig. 10".

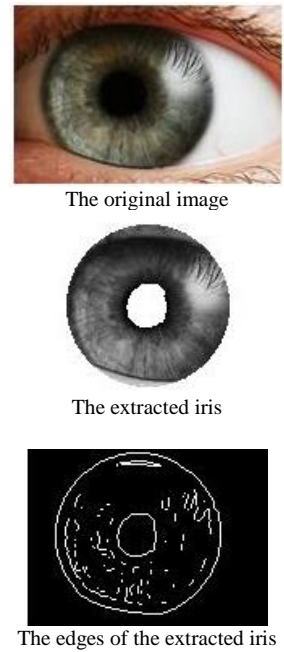

The original image

The extracted iris

The edges of the extracted iris

Fig. (7), Using Entropy-based texture technique; the original image is illustrated with the extracted iris, as we show an optional result for the edges of the iris region.

IV. THE VIGILANT DRIVER:

Both iris segmentation techniques, already discussed in the third paragraph, give us a so high reliable performance. To prove this fact we have tested both of them with 315 different face images, taken from different angles. The percentage of correct pupil detection and iris segmentation for any given image using the Entropy-based texture description is 95.23%. Whereas, for the Gabor-based technique it is 99.04%. Therefore, we will consider the second technique as the main part of our system to recognize the level of driver vigilance depending on two measured values; (f) the frequency of eye glances, and (T) the interval during which the driver's eye is closed.

According to these values; and depending on the fuzzy sets defined as shown in "fig.11", we can distinguish a set of different behaviors of the vehicle driver. For the first measured value (f) we have suggested two trapezoidal fuzzy sets (Normal and high), the ranges of frequencies (glance per minute) for each fuzzy set are given as illustrated in "fig. 11". While for the interval of a closed eye (T) we can distinguish four fuzzy sets (Very short, short, long, and very long), the membership functions of the output called (Unconsciousness) are also illustrated in the same figure.

The reader may think that an additional third fuzzy set that should be assigned to (f), could be called (Low) and may look like dropped from this design. But in fact, this fuzzy set is logically equal to the fuzzy set (Very Long) of the second variable (T), i.e. when T is very long (f) should be near to zero.

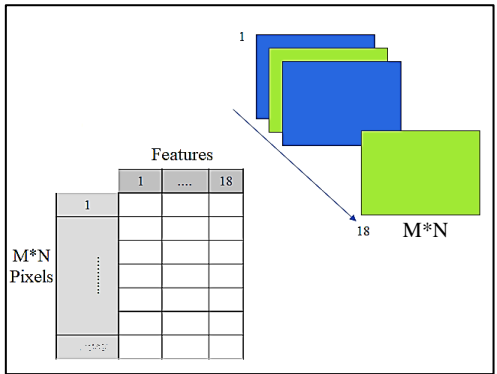

Fig. (8), A number of successive images of extracted Gabor features plus two layers of spatial features in X and Y. All images are converted into one unique array which is an expected form by the PCA procedure.

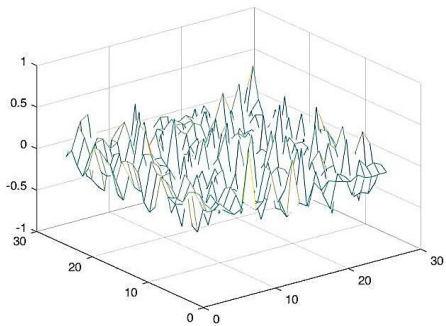

Fig. (9), The principal component coefficients extracted by PCA.

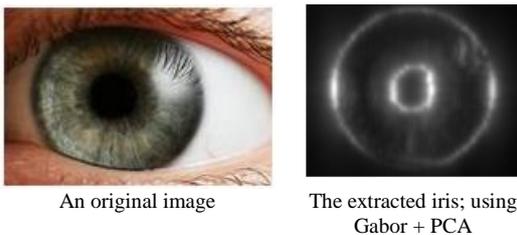

An original image     The extracted iris; using Gabor + PCA

Fig. (10), The extracted iris region using the Gabor-based technique with the PCA procedure.

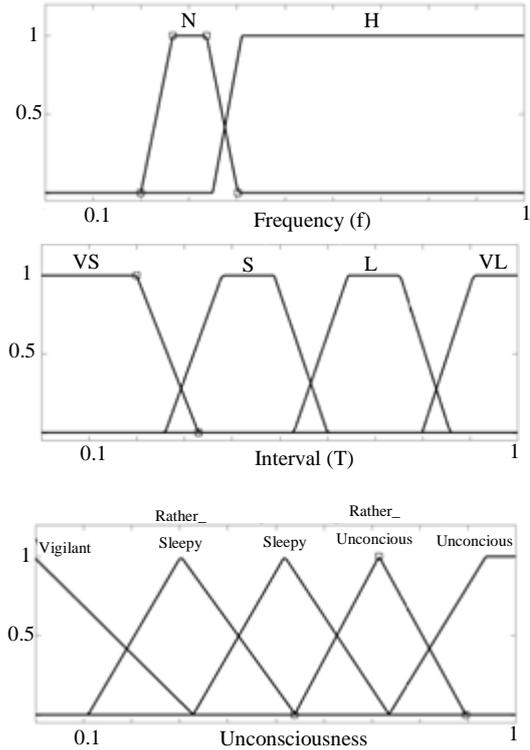

Fig. 11, The suggested fuzzy sets for input and output variables

According to the upper mentioned description of the fuzzy system, the following table could be established to distinguish six fuzzy rules as shown in table 1 and "fig. 12". While, for an additional and unique case, when (T) is "Very Long" or (f) is zero; the system should directly conclude that the driver is completely unconscious, then the driver should be alerted. 3D model for the control surface generated by this fuzzy system is shown if "fig. 13".

TABLE I.     6 DIFFERENT LEVELS OF DRIVER VIGILANCE

| T / f | VS | S | L |
|---|---|---|---|
| N | Vigilant | Rather Sleepy | Sleepy |
| H | Rather Sleepy | Sleepy | Rather Unconscious |

## V. CONCLUSION AND PERSPECTIVES

In this paper, we have presented a fuzzy logic-based system to distinguish the behavior of a vigilant driver. The main measured values which are the frequency of the eye glance and the interval of eye closing, depend directly on the accuracy of our approaches for achieving pupil detection and iris segmentation. Therefore, we have presented two different techniques of iris segmentation as we have corrected the location of the detected pupil for any given image. Face recognition and human eye localization were already discussed in some already published papers.

In the near future, more important tests and practical experiments should be achieved to perform this embedded system as a final product. As we intend to use an infrared camera to get the temperature of the car driver taking into account the temperature of the internal cabin. This corporal measurement seems promising in order to give a more robust description of the driver's vigilance case.

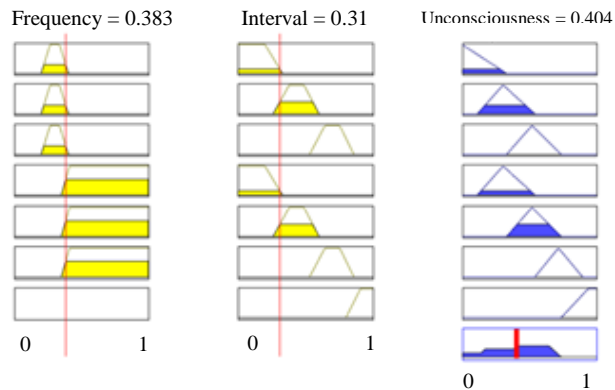

Fig. 12, An example of MATLAB-based simulation for the 7 fuzzy rules suggested here.

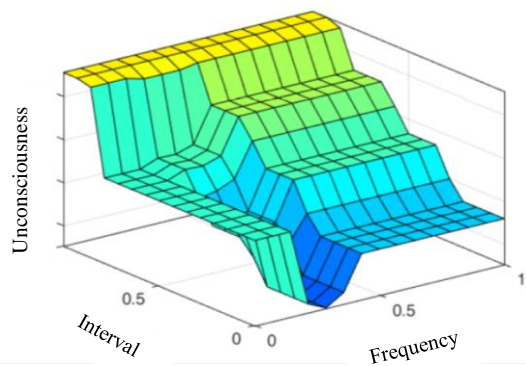

Fig. 13, The 3D model for the control surface generated by the fuzzy system